\documentclass[journal]{IEEEtran}
\usepackage{graphicx}
\usepackage{amsmath,amssymb}
\usepackage{color}
\usepackage{graphicx}
\usepackage{algorithm}
\usepackage{algorithmic}
\usepackage{subfig}
\usepackage{multirow}
\usepackage{url}
\usepackage{cite}
\usepackage{bm}
\usepackage{array}
\usepackage[american]{babel}
\makeatletter
\adddialect\l@AMERICAN\l@american
\makeatother
\usepackage{microtype}
\usepackage{booktabs}
\usepackage[T1]{fontenc}

\newcommand{\x}{{\bm x}}

\newcommand{\T}{{\top}}
\newcommand{\tr}{{\rm tr}}

\newcommand{\eg}{\textit{e.g.}}
\newcommand{\ie}{\textit{i.e.}}
\newcommand{\wrt}{\textit{w.r.t.}}

\begin{document}
\title{Structure-Aware Consensus Network on Graphs with Few Labeled Nodes}
\author{
Shuaike Xu, Xiaolin Zhang, Peng Zhang, and Kun Zhan
\thanks{Manuscript received July 2, 2024; revised November 20, 2024.}
\thanks{Shuaike Xu and Kun Zhan are with the School of Information Science and Engineering, Lanzhou University, Lanzhou, Gansu 730000, China. (e-mail: kzhan@lzu.edu.cn)}
\thanks{Xiaolin Zhang is with College of Electrical Engineering and Automation,  Shandong University of Science and Technology, Qingdao, China}
\thanks{Peng Zhang are with College of Computer Science and Engineering, Shandong University of Science and Technology, Qingdao, China.}
\thanks{This work was supported by the National Natural Science Foundation of China under Grant No.~6217610.}
}
\markboth{IEEE Transactions on Cybernetics,~Vol.~14, No.~8, August~2026}
{Xu \MakeLowercase{\textit{et al.}}: SACN}
\maketitle
\begin{abstract}
Graph node classification with few labeled nodes presents significant challenges due to limited supervision. Conventional methods often exploit the graph in a transductive learning manner. They fail to effectively utilize the abundant unlabeled data and the structural information inherent in graphs. To address these issues, we introduce a Structure-Aware Consensus Network (SACN) from three perspectives. Firstly, SACN leverages a novel structure-aware consensus learning strategy between two strongly augmented views. The proposed strategy can fully exploit the potentially useful information of the unlabeled nodes and the structural information of the entire graph. Secondly, SACN uniquely integrates the graph's structural information to achieve strong-to-strong consensus learning, improving the utilization of unlabeled data while maintaining multiview learning. Thirdly, unlike two-branch graph neural network-based methods, SACN is designed for multiview feature learning within a single-branch architecture. Furthermore, a class-aware pseudolabel selection strategy helps address class imbalance and  achieve effective weak-to-strong supervision. Extensive experiments on three benchmark datasets demonstrate SACN's superior performance in node classification tasks, particularly at very low label rates, outperforming state-of-the-art methods while maintaining computational simplicity.
\end{abstract}

\section{Introduction}\label{Intro}
Due to the high cost of labeling graph data, it is essential to develop semi-supervised graph node classification methods that perform with only a few labeled samples. Graph node classification becomes significantly challenging when labeled data is very limited. 

\begin{figure}[th]
\centering
\includegraphics[width=0.48\textwidth]{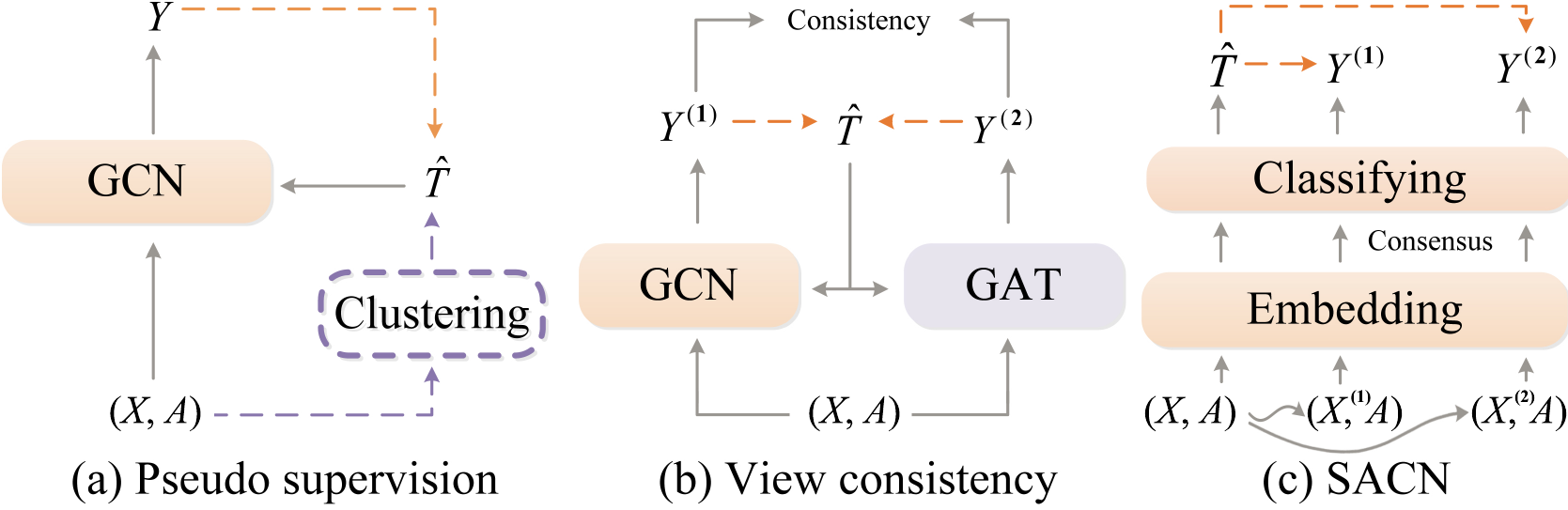}\\
\caption{Different self-supervised strategies. Attributed feature matrix $X$ and graph structure $A$ to the model, output prediction representation $Y$. (a) Previous simple pseudo supervision strategy. (b) View-consistency strategy. (c) the proposed SACN mainly uses a structure-aware consensus objective between strongly augmented views. Furthermore, we use a weak-to-strong supervision between $\hat{T}$ and $Y^{(v)}$. We use a class-aware strategy to generate pseudolabels $\hat{T}$, which is used to supervise the predictive representations ${Y}^{(1)}$ and ${Y}^{(2)}$.}\label{fig01}
\end{figure}

Methods for graph node classification with few labeled nodes are categorized into three types, as shown in Fig.~\ref{fig01}. 
Early algorithms generate pseudolabels in two different ways. One method involves simply self-train by thresholding prediction scores~\cite{lhw:AAAI2018} to roughly estimate the pseudolabels which is represented by the yellow dashed line at the top of Fig.~\ref{fig01}(a). 
The other method utilizes auxiliary tasks~\cite{gcl:mutual2020} to cluster the nodes to produce pesudolabels which is indicated by the purple dashed line at the bottom of Fig.~\ref{fig01}(a). To ensure reliability of pseudolabels, multiview voting approaches that estimate pseudolabels by voting across different views are developed, as shown in Fig.~\ref{fig01}(b). 
For instance, the view consistency heterogeneous network establishes a heterogeneous structure and votes among predictions~\cite{liao-et-al:View-Consis}. 

However, these approaches face two major issues: 1) simply using an auxiliary task often fails to produce reliable pseudolabels; 2) the view consistency method~\cite{liao-et-al:View-Consis} requires extra networks for cross-view learning requiring unnecessary computational resources. In this paper, we think the key to improve the classification performance is to fully utilize the information from the abundant unlabeled data. 
The structural information inherent in graphs give cues to mine these useful information from the unlabeled.

We propose a Structure-Aware Consensus Network (SACN) to fulfill this purpose and to boost the classification performance from three perspectives. 
Firstly, SACN not only maintain a single-branch network, but also leverages unlabeled data through cross-view supervision without extra branches. Secondly, as shown in Fig.~\ref{fig01}(c), We introduce a structure-aware consensus learning objective for graphs of two strongly augmented views, \ie, $X^{(1)}$ and $X^{(2)}$. 
This objective is to maximize the cross-view consensus for the same class and disentangle irrelevant classes, inspired by the approach used in unsupervised learning methods~\cite{Harold:varsrelation,zhang-et-al:CCA-SSG}. SACN is different from these methods~\cite{Harold:varsrelation,zhang-et-al:CCA-SSG} which optimize the distance between different instances, the proposed objective maximize the consensus for the same class-prediction across the two augmented views. Our approach has two main advantages: 1) The consensus information between dual-view latent features with edge constraints is easy to compute, introducing only marginal computational burden; 2) Maximizing the same objective across different feature channels enhances the robustness of features, especially across different views. Our approach effectively leverages the self-supervised nature of consensus learning objective to further enhance the learning of relevant information from graphs. Thirdly, we propose to utilize pseudolabels from the weakly augmented view to supervise the prediction results of the two strongly augmented views in SACN, which can further improve classification accuracy. On the node classification task, our approach achieves results comparable to the state-of-the-art, showcasing the value of graph structure information.
 
We propose to optimize the SACN model with three derived objectives. We apply a parallel random augmentation process to the inputs then construct the consensus objective between the learned features. This loss innovatively combines structure-aware consensus learning objective with specific structural information, to correlate features from adjacent nodes and decorrelate features from different classes.
Furthermore, for weak-to-strong supervision, a class-aware pseudolabel generation strategy is proposed carefully on the weakly augmented view to supervise the two strongly augmented predictions. In practice, we do label selections on every individual class instead of the whole set, to avoid low accuracy on imbalanced classes. Extensive experimental results on the node classification task show that under simple network design, SACN achieves comparable performance of state-of-the-art methods on most benchmarks with limited labeled nodes. We also conduct a comprehensive ablation study to verify the effectiveness of key components in our approach.

The contributions are summarized as follows:
\begin{itemize}

  \item We introduce a strong-to-strong structure-aware consensus learning objective for unlabeled data in semi-supervised node classification. 
  \item  
  To the best utilization of unlabeled data, we propose to use the multiview predictions on the weakly augmented view as pseudolabels to supervise the prediction of two strongly augmented views. 
  \item Under the weak-to-strong supervision, a novel class-aware pseudolabel generation strategy is introduced to alleviate the impact of the imbalanced distribution. 
\end{itemize}

The rest of the paper is organised as follows: We introduce some related work in Section \ref{RW}. We propose SACN to address semi-supervised node classification with imbalanced data with very low label rates In~Section \ref{sec_method}. 
In Section \ref{Experiments}, we conduct experiments to demonstrate the effectiveness of SACN. In Section \ref{Conclusions}, we present conclusions of the paper.
\section{Related Work}\label{RW}
\subsection{Contrastive Learning on Graphs}
As a common sense, humans always observe the same object from completely different perspectives. Though the descriptions of one object vary from people, these descriptions always refer to the same one, \ie, those descriptions from different perspectives share the same semantics. Making use of this multi-view consistency, contrastive methods often exchange information cross various views or branches by making them consistent, to receive accurate predictions of samples. Multiview is realized by transformation or perturbation, while multi-branch is realized by introducing heterogeneous network.

With the strong motivation, contrastive methods~\cite{hjelm-et-al:learning} have been shown to be effective for unsupervised learning in computer vision, which have been adapted to graphs. For instance, inspired by the mutual information maximization view~\cite{hjelm-et-al:learning}, DGI~\cite{DGI} proposed unsupervised schemes for nodes. 
MVGRL~\cite{hk:contramulti} promotes CMC~\cite{tdp:contracoding} to graph structure by introducing diffusion~\cite{jss:diffusion} to graph to generate another view on it. While GRACE~\cite{zhu-et-al:deep2020} follows SimCLR~\cite{chen-er-al:FrameforCL} to learn node representatio. M-ILBO~\cite{ENSmymm2023} estimates the entropy of the dataset in a graph contrastive learning manner.
\subsection{Pseudo Supervision Strategy}
In the field of semi-supervised learning, Lee~\cite{lee:Pseudo-Label} first proposed the pseudolabel method that adds labeled data and unlabeled data into the training process. For unlabeled data, the label corresponding to the maximum prediction probability which is selected during each weight update, and the obtained pseudolabels are regarded as supervised information to expand the originally insufficient label set. Regarding the existent pseudolabeling strategies, there are three main categories: (a) Self-training strategy~\cite{berthelot-et-al:2019mixmatch,sohn-et-al:2020fixmatch,lhw:AAAI2018}. This simplest strategy involves setting a threshold on the model's predicted probabilities then binarize the higher ones to generate pseudolabels. However, this approach may result in noisy pseudolabels. (b) Auxiliary clustering-based strategy~\cite{slz:multi2019,gcl:mutual2020}. To guide label assignment, this kind of approaches utilize clustering algorithms based on similarity in feature space. As name suggests, this strategy aims to group similar nodes together and assign them the same pseudolabel. (c) consistency regularization strategy~\cite{olr:TADAM,beyer-et-al:S4L,liao-et-al:View-Consis}. These methods estimate pseudolabels by seeking consistency among multiple views. This is achieved by combining predictions from different models or different views. This one aims to improve label quality and reduce noise.

However, the mentioned previous methods mostly compare predictive probabilities on the whole data, ignoring the impact of imbalanced distribution. The class-aware pseudolabel strategy adopted in this paper is friendly to the long-tail distribution. By comparing the prediction probabilities on each class, we obtain labels with high confidence.
\subsection{Data Augmentation Strategy}
To address the lack of labels in graph node classification, many data augmentation strategies is proposed~\cite{you-et-al:GraphCl}. The basic idea of data augmentation is to create novel and realistic data by applying certain sensible transformations. In addition, the augmentation process does not compromise semantic labels. The graph contrastive learning methods~\cite{hk:contramulti,peng-et-al:MutualInfoMax,zhu-et-al:deep2020} generate different views by disrupting the original graph structure. For instance, GRACE~\cite{zhu-et-al:deep2020} generates views with different types of graph augmentations and learns the node representation by maximizing the consistency of the node representation between views. The current data augmentation methods on graphs are divided into four types~\cite{you-et-al:GraphCl}: node masking~\cite{fang-et-al:dropmessage,hk:bootstrapped2021}, edge masking~\cite{2021robust,luo-et-al:edgedrop}, feature masking~\cite{feng-et-al:grand,zhu-et-al:gca,you-et-al:gcla} and subgraph partitioning~\cite{yang-et-al:DSGC,liu-et-al:subgraph}. Inspired by~\cite{thakoor-et-al:bootstrapped2021,zhu-et-al:gca,hu-et-al:PreGNN}, We perform feature masking by setting certain dimensions of the node attributes to zero, ensuring that the graph structure remains intact. In order to highlight the effect of the proposed new objective function integrating neighborhood information, we just use the feature mask strategy. 
\section{Structure-Aware Consensus Network}\label{sec_method}
\subsection{Preliminaries}
Let $G=(X,A)$ denote a graph , where $X=[\x_1,\x_2,\ldots,\x_n]^\T \in\mathbb{R}^{n \times m}$ denotes the feature matrix of $n=l+u$ nodes, where $l$ is the number of labeled nodes and $u$ is the number of unlabeled nodes. $A=[a_{ij}]\in \{ 0, 1\}^{n \times n}$ is the edge affinity matrix and  $a_{ij}$ denotes the pairwise connection between the $i$-th node and the $j$-th node. $a_{ij}=1$ for two adjacent nodes and $a_{ij}=0$ otherwise.

Specifically, let ${X}^{(1)}$ and ${X}^{(2)}$ denote two randomly augmented views of the input feature. By feeding them into a graph neural network $f(\cdot|\theta)$, where $\theta$ is the parameter of the neural network, we obtain their corresponding latent feature representations ${Z}^{(1)}\in \mathbb{R}^{n \times d}$, ${Z}^{(2)} \in \mathbb{R}^{n \times d}$,
\begin{equation}\label{representations}
 {Z}^{(1)}=f({X}^{(1)},A|\theta), {Z}^{(2)} = f({X}^{(2)},A|\theta) .
\end{equation}
where $d$ is the feature dimension of each node.

Latent features from different views of the same target ought to be consistent with the optimization process.
Thus, the graph neural network can distinguish between different nodes and structures for the classification purpose.
By augmenting the learning process with structural information,  features of the unlabeled nodes are more likely to be accurately mined.
Inspired by~\cite{Harold:varsrelation,pmlr-v28-andrew13,wang-et-al:deep2015,zhang-et-al:CCA-SSG}, we employ an channel-wise optimization approach to boost the robustness of node features. The correlation is computed and formulated by
\begin{equation}\label{CCA}
\begin{aligned}
\max_{\theta}\,& \tr\Bigl(\bigl(Z^{(1)}\bigr)^\top Z^{(2)}\Bigr) \\
{\rm s.t.}\,& \bigl(Z^{(1)}\bigr)^\top Z^{(1)} =\bigl(Z^{(2)}\bigr)^\top Z^{(2)} =I ,
\end{aligned}
\end{equation}
where $\tr(\cdot)$ denotes the trace operation, and $I$ is an identity matrix. Eq.~\eqref{CCA} maximizes correlation between features from the two different views while the corresponding constraints decorrelate the inter-dimension dependence within each view.

Compared to contrastive learning, this consensus learning is different in the way processing information. In particular, if the prediction involves $n$ points with $c$ channels, contrastive learning primarily evaluates the point-to-point consistency. It assesses whether two points constitute positive pairs, while consensus learning focuses on comparing entire sets of points within one channel from one view with those from the other view. Consensus learning aims to maximize the cross-view agreement for the same class and disentangle irrelevant classes. In this paper, consensus learning is achieved through mutual learning among nodes across augmented views, it integrates the neighborhood information unique to graph structure. 

\begin{figure}[!tp]
  \centering
  \includegraphics[width=0.3\textwidth]{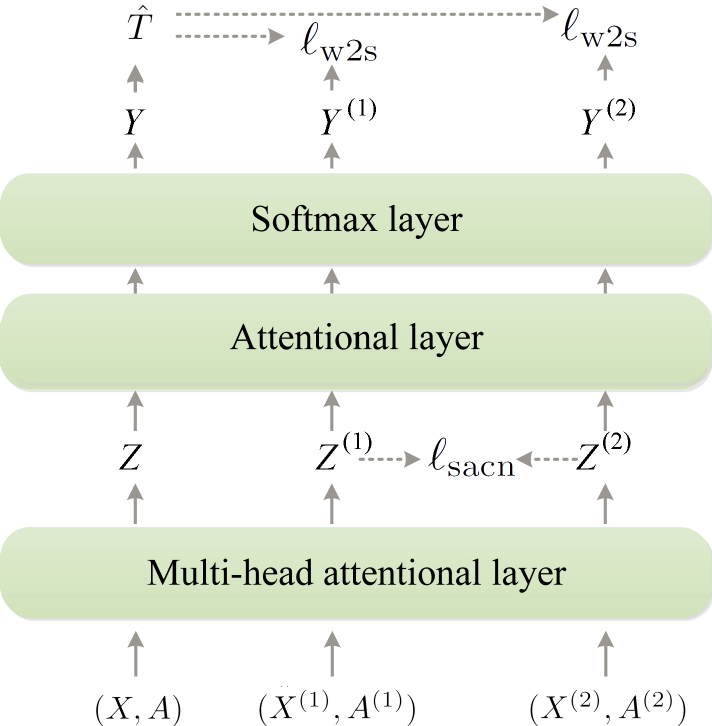}\\
  \caption{Structure-aware consensus network architecture. SACN is constructed on a two-layer graph attention network, consists of two augmented views and one normal view.}\label{fig02}
\end{figure}
\subsection{Structure-Aware Consensus Objective}
As shown in Figs.~\ref{fig01} and \ref{fig02}, the primary purpose of the first layer is to learn latent embedding representations, while the second layer is dedicated to the downstream classification task. With this in mind, we employ a view-consensus learning approach with a dual-augmentation strategy for the output of the first layer, while the design of the second layer resembles that of a downstream task. In details, SACN feds two parallelly augmented views to encoder to obtain the corresponding features $Z^{(1)}$ and $Z^{(2)}$. Since SACN shares parameters across views, it is appropriate to build consensus constraints between the resulting encoded features. Firstly, we try to utilize the above Eq.~\eqref{CCA} which performs maximization of the trace on the orthogonal of the two representations to achieve consistency.
The trace operation is expressed in matrix form, and Eq.~\eqref{trace} shows the element form at each position.
\begin{equation}\label{trace}
\tr\Bigl(\bigl({Z^{(1)}}\bigr)^\T {Z}^{(2)} \Bigr)
= \sum_{i=1}^n\bigl({\bm {z}^{(1)}_{i}}\bigr)^{\T}  \bm {z}^{(2)}_{i}\,.
\end{equation}

It seems obviously that the above method maximizing traces enforces the same nodes on different views to be aligned with each other. However, we present an interesting insight, that is, in addition to the inevitable cross-view consensus of the node itself, the graph structure also implies associative relationships. It makes an intuitive sense that applying the cross-view consensus constraint to more nodes that have relationships works better. 

For the motivation of the proposed SACN approach, we consider two basic rules: 1) the features between related nodes are optimized to be close, and thus to aid the classification of the unlabeled nodes;
2) the features of nodes from different categories are learned apart with the training process.
Therefore, the learned features of the nodes result in improving the performance on node classification while reducing the sensitivity.

Practically, to satisfy the first rule, the most straightforward approach is to minimize the sum of the elements on the diagonal after the two representations are orthogonal as shown in Fig.~\ref{fig02}. However, previous contrastive methods doing such operations only correlates the same node from the two views, without considering the actual adjacency of graph, which violates the first rule. To improve, we add the valuable neighborhood information, which gives the last two terms of $\ell_{\rm cor\_x_i}$ in Fig.~\ref{fig02}.
Specifically, we employ the adjacency matrix $A=[a_{ij}]$ to formalize the correlation of the adjacent nodes from the two differently augmented views: 
\begin{equation}\label{inv_loss}
 \ell_{\rm cor} =  - \sum_{i=1}^n\sum_{j=1}^na_{ij}\bigl({\bm {z}^{(1)}_{i}}\bigr)^{\T}  \bm {z}^{(2)}_{j}.
\end{equation}
The features of the adjacent nodes have larger weights, ~\ie, $a_{ij}=1$, while the disconnected nodes have smaller weights, ~\ie, $a_{ij}=0$. It is obvious that minimizing this $\ell_{\rm cor}$ equals to maximizing the correlation between adjacent nodes from two views. This optimization brings features of adjacent nodes correctly to be close, while the features of disconnected nodes do not be wrongly brought to be similar.

It can be seen from the above Eq.~\eqref{inv_loss} that the distinction between SACN and other CCA-like losses is evident. Traditional CCA methods often involve intricate matrix decomposition or optimization that become prohibitively expensive for large graphs. In contrast, approaches based on the inner product like SACN avoid these complexities, making it more accessible and practical for a wider range of applications. In Eq.~\eqref{inv_loss}, the form of inner-product captures the essence of the relationship between two adjacent nodes, thereby preserving the critical information necessary for correlation analysis. Especially when dealing with sparse graphs, the advantage of this loss is verified. In sparse graphs, the majority of the nodes are not directly connected, resulting in a significant reduction in the number of inner-product computations required. By focusing solely on the inner-product computation integrating adjacency information, it bypasses the need of traditional CCA for complex matrix operations or extensive feature extraction. This $\ell_{\rm cor}$ not only maintains the integrity of the correlation analysis reducing the computational burden but also combines neighborhood relevance. 

To satisfy the second rule, we force the features across different dimensions to be distinct as feature dimensions are directly corresponding to the probabilities of the predicted categories. Then, $Z^{(1)}$ and $Z^{(2)}$ are $z$-score normalized with a zero mean and a standard deviation. We do this batch normalization in SACN to benefit comparison between dual-view features.
Also, distinct features of different classes reduce the misclassification chances.
Practically, we use the constraint of Eq.~\eqref{CCA} on the feature of each view as shown in Eq.~\eqref{dec_loss}.
\begin{equation}
\ell_{\rm de} = \left\|\bigl({Z}^{(1)}\bigr)^\T{Z}^{(1)} - I\right\|^{2}_{\rm F}+\left\|\bigl({Z}^{(2)}\bigr)^\T{Z}^{(2)} - I\right\|^{2}_{\rm F} \,.\label{dec_loss}
\end{equation}

Minimizing $\ell_{\rm de}$ is actually pushing the off-diagonal elements of the covariance matrix of ${ Z}^{(1)}$ and ${ Z}^{(2)}$ close to zero. 
Here, an interesting insight is that Eq.~\eqref{dec_loss} is actually a measure to solve the imbalance of class distribution. By constraining the feature matrix approaching to the identity matrix across dimensions, it limit every node to only belong to one class on the graph. 
Specifically, the matrix ${Z}^{(1)}$ and ${Z}^{(2)}$ are already normalized on each column,~\ie, normalize on each dimension vector. 
It essentially adds constraints on the number of nodes in each class, and reduces the impacts of the imbalance issue. 

We combine the two above losses, and it comes to the following structure-aware consensus objective:
\begin{align}
& \ell_{\rm sacn} = \ell_{\rm cor}+\lambda\ell_{\rm de},
\label{aug_loss}
\end{align}
where $\lambda$ is a trade-off parameter.
\subsection{Weak-to-Strong Supervision}
We introduce a weakly augmented view to supervise the two strongly augmented views. Here we use `weakly' to distinguish it from the other two strongly augmentations, for there is not any augmentation at all on the weak view.

On the bottom of Fig.~\ref{fig02}, the input pass through the multi-head attentional layer to learn weakly augmented representation ${Z}$, then ${Z}$ pass through the further attentional layer and a softmax layer to obtain the corresponding soft predictions, ${Y} = [y_{ij}]$. On the two strongly augmented views, soft predictions $Y^{(1)}$ and $Y^{(2)}$ are generated in the same way.

Unlabeled samples lack true labels, we treat the predictions on them (\eg, soft predictions ${Y} = [y_{ij}]$) as the pseudolabels. The strategy of using the prediction of unlabeled samples as its own supervision information is entropy regularization~\cite{luan-et-al:multiscale}. Differently, we select high confidence ones in soft predictions ${Y}$ on the weakly augmented view, turn the selected soft predictions to hard pseudolabels $\hat t=[{t_{ij}}]$, then use $\hat t$ to supervise other two strongly augmented views instead of the weakly augmented view itself. Notably, to address the serious class imbalance issue that graph data typically faces, we introduce a novel class-balanced pseudolabel generation strategy. For the reason why we choose high confidence scores,  high confidence usually indicates the high accuracy of the obtained pseudolabels. The obtained $\hat t$ is used to guide the strongly augmented results. We construct weak-to-strong supervision in the form of cross-entropy between pseudolabels $\hat t$ and the two strongly augmented predictions,~\ie, ${Y}^{(1)}$ and ${Y}^{(2)}$, separately:

We draw on the direct GNN approaches which usually maximize a likelihood function. The likelihood function is defined through the classification conditional probability,
\begin{equation}\label{conditional}
p(\omega_j|\bm{g}_i)=y_{ij},
\end{equation}
where $y_{ij}$ denotes the predicted probabilities of class $j\in\{1,2,\ldots,k\}$ \wrt~the $i$-th input node feature $\bm{g}_{i}$, $k\,$ is the number of the classes. $p(\omega_j|\bm{g}_i)$ is obtained by an output of a ${\rm softmax}$ layer, while the dimension of the ${\rm softmax}$ layer equals to $k\,$, and $\omega_j$ denotes a certain class. 

The maximization of likelihood is completed using the 1-of-$k$ one-hot coding strategy. To be specific, for a certain node $\bm{g}_i$, the corresponding target vector $\bm t_i$ is a binary vector with all elements zeros except for one element $j$. In $\bm t_i$, element $j$ equals one, denoting that the $i$-th node belongs to class $\omega_j$.

Based on Eq.~\eqref{conditional}, the likelihood function is given by
\begin{equation}\label{likelihood}
p(T|X,A)=\prod_{i=1}^l\prod_{j=1}^ky_{ij}^{t_{ij}},
\end{equation}
where $T$ is an $l\times k$ matrix of target variables, whose elements are denoted as $t_{ij}\,$.
\begin{equation}\label{ta}
  \ell_{\rm w2s}=-\sum_{i=1}^u\sum_{j=1}^k \left(\hat t_{ij}\ln {y}^{(1)}_{ij}+ \hat t_{ij}\ln {y}^{(2)}_{ij}\right)\ ,
  \end{equation}
where $u$ denotes the number of high confidence predictions.
\begin{algorithm}[!t]
\caption{The SACN algorithm.}\label{algorithm}
\begin{algorithmic}[1]
\STATE \textbf{Input}: Dataset $X$ and its graph $A$
\STATE \textbf{Output}: $p(\omega_j|\x_i)$, $\forall\,i,j$
\STATE \textbf{Initialization:} $epoch, epoch_{\max}$ and $epoch_{\rm pre}$
\WHILE{$epoch\leq {epoch_{\max}}$}
    \IF {$epoch\leq {epoch_{\rm pre}}$}
        \STATE Calculate the loss using Eq.~\eqref{loss_sup}\,.
        \STATE Update parameters by backward propagation.
    \ELSE
    \STATE Calculate the loss using Eq.~\eqref{loss_overall}\,.
        \STATE Update parameters by backward propagation.
        \IF {Converges}
            \STATE Break.
        \ENDIF
    \ENDIF
    \STATE $epoch = epoch + 1.$
\ENDWHILE
\end{algorithmic}
\end{algorithm}
Taking the negative logarithm of Eq.~\eqref{likelihood}, the loss function is derived in the form of the cross-entropy,
\begin{equation}
\ell_{\rm sup}=-\ln p(T|X,A)=-\sum_{i=1}^l\sum_{j=1}^kt_{ij}\ln y_{ij}\ .\label{loss_ce}
\end{equation}
Minimizing this loss is equivalent to maximizing the likelihood in Eq.~\eqref{likelihood}, following this general idea, we adopt cross-entropy function in weak-to-strong supervision of Eq.~\eqref{ta}.

For detailed generation of the hard pseudolabel $\hat t_{ij}$, it is converted by one-hot encoding operation based on $y_{ij}$. Concretely, the soft prediction result $y_{ij}$ is obtained by softmax, then we assign the predicted class with the highest probability to the hardmax $\hat t_{ij}$.

Notably, the process of selecting high predicted confidence pseudolabels on each view is carried out for each class, namely the class-aware strategy. Most predicting methods set the same confidence threshold for all classes, which is very unfriendly for imbalanced datasets. 
In contrast, class-aware strategy set the unique threshold varying with classes, which means the selection of high confidence predicted soft labels is carried out for each class separately, so as to alleviate the problem caused by imbalanced distribution.
\subsection{Algorithm}
Fig.~\ref{fig02} depicts the overall structure of the proposed SACN. 
SACN is constructed by two graph attentional layers~\cite{graph2018}. Each layer aggregates local features with self-attention coefficients. For efficiency, the first layer is designed with multiple heads and primarily focuses on learning representations. To guide this process, we introduce the SACN loss Eq.~\eqref{aug_loss} to operate on the output of the first attention layer, it composed of Eq.~\eqref{inv_loss} and Eq.~\eqref{dec_loss}. In the middle of the Fig.~\ref{fig02}, it is worth emphasizing that the $\ell_{\rm cor}$ of SACN formed in Eq.~\eqref{inv_loss} is different from the general contrastive learning loss. We take into account the unique structural information of the graph to ensure the maximum correlation between nodes to a greater extent. The representations of multiple heads are of concatenation then performed dropout. This strategy renders SACN obtain stable predictions. After dropout, the latent feature on the weakly augmented view,~\ie, $Z$, passes through the second graph attention layer and the ${\rm softmax}$ layer to obtain the final soft prediction, $Y=[y_{ij}]$. This implies that the second attention layer is primarily utilized for the downstream task, \ie, graph node classification.

In the first stage, we use labeled nodes to pre-train network to achieve a certain initialization effect and improve the reliability of pseudolabels, with the SACN loss,
\begin{equation}\label{loss_sup}
\ell_{\rm one}= \ell_{\rm sup}+\alpha_{1}\ell_{\rm sacn}
\end{equation}
where $\alpha_{1}$ is a hyperparameter.

In the second stage, firstly, the augmented graphs after random mask are fed into the corresponding views to generate two latent features. 
Then, we apply a consistency constraint of the form Eq.~\eqref{aug_loss} between them. At the same time, the pseudolabels generated on the weakly augmented view are used to supervise the soft predictions from the two strongly augmented views respectively in the form of Eq.~\eqref{ta}. 
 
Then, the second stage loss is given by
\begin{equation}\label{loss_overall}
\ell_{\rm two} = \ell_{\rm sup}+\alpha_{1}\ell_{\rm sacn} +\alpha_{2} \ell_{\rm w2s} ,
\end{equation}
where $\alpha_{2}$ is a hyperparameter.

We summarize the SACN algorithm in Algorithm~\ref{algorithm}.

\section{Experiment}\label{Experiments}
We conduct experiments on three popular benchmarks ~\ie, Cora, Citeseer, and PubMed~\cite{sen-at-al:data} with different label rates to demonstrate the effectiveness of the proposed SACN method. Ablation experiments are conducted to illustrate the usefulness of the designed components.
The detailed information of the three datasets are summarized in Table~\ref{dataset}.
\begin{table}[!ht]
  \caption{Datasets statistics.}
  \label{dataset}
  \centering
\begin{tabular*}{0.42\textwidth}{@{\extracolsep{\fill}\,}l|llll}
    \toprule
    Dataset & Vertices  & Edges & Features  &Classes  \\
    \midrule
    Cora & 2708 & 5429 & 1433 & 7 \\
    Citeseer & 3327 & 4732 & 3703 &6  \\
    PubMed & 19717 & 44338 & 500 & 3 \\
    \bottomrule
  \end{tabular*}
\end{table}

\begin{table*}[ht]
\centering
\caption{Accuracy With Validation (\textbf{Bold} values indicate the best).}\label{with}
\begin{tabular*}{\textwidth}{@{\extracolsep{\fill}\quad}l|ccc|cc|ccc}
\toprule
Dataset         &\multicolumn{3}{c|}{Cora}                           &\multicolumn{2}{c|}{Citeseer}   &\multicolumn{3}{c}{PubMed} \\
Label rate        &0.5\%            & 1\%           & 3\%           &0.5\%          & 1\%           &0.03\%         & 0.05\%        & 0.1\% \\
\midrule
Chebyshev         &33.9             &44.2           &62.1           &45.3           &59.4           &45.3           &48.2           &55.2   \\
GCN-FP            &50.5             &59.6           &71.7           &43.9           &54.3           &56.2           &63.2           &70.3   \\
GraphSAGE         &37.5             &49.0           &64.2           &33.8           &51.0           &45.4           &53.0           &65.4   \\
GAT               &41.4             &48.6           &56.8           &38.2           &46.5           &50.9           &50.4           &59.6   \\
GCN               &50.9             &62.3           &76.5           &43.6           &55.3           &57.9           &64.6           &73.0   \\
LNet              &58.1             &66.1           &77.3           &53.2           &61.3           &60.4           &68.8           &73.4   \\
AdaLNet           &60.8             &67.5           &77.7           &53.8           &63.3           &61.0           &66.0           &72.8   \\
Linear Snowball   &70.0             &73.1           &81.0           &59.4           &65.9           &68.1           &70.0           &73.8   \\
Snowball          &73.0             &76.8           &80.7           &62.1           &64.2           &70.8           &73.2           &76.5   \\
Truncated Krylov  &73.9             &77.4           &82.2           &63.7           &68.4         &71.1           &72.9           &75.7   \\
VCHN              &{74.9}           &{77.8}         &{83.1}         &{65.6}         &\textbf{70.1}          &{71.8}         &{74.3}         &{76.8} \\
\midrule
\textbf{SACN}    &\textbf{76.2}  &\textbf{79.2}  &\textbf{83.6} &\textbf{67.0}  &{69.5}  &\textbf{72.2}  &\textbf{74.4}     &\textbf{77.4}  \\
\bottomrule
\end{tabular*}
\end{table*}

\begin{table*}[ht]
\centering
\caption{Accuracy Without Validation (\textbf{Bold} values indicate the best).}
\label{without}
\begin{tabular*}{\textwidth}{@{\extracolsep{\fill}\quad}l|ccc|cc|ccc}
\toprule
Dataset         &\multicolumn{3}{c|}{Cora}       &\multicolumn{2}{c|}{Citeseer}     &\multicolumn{3}{c}{PubMed} \\
Label rate      &0.5\%     &1\%        &3\%         &0.5\%  & 1\%                     &0.03\%  &0.05\%   &0.1\% \\
\midrule
LP              &56.4      &65.4       &67.5        &34.8   &40.2                      &61.4   &66.4   &65.4   \\
Chebyshev       &38.0      &62.4       &70.8        &31.7   &42.8                      &40.4   &47.3   &51.2   \\
Co-training     &56.6      &73.5       &75.9        &47.3   &55.7                      &62.2   &68.3   &72.7   \\
Self-training   &53.7      &73.8       &77.2        &43.3   &58.1                      &51.9   &58.7   &66.8   \\
Union           &58.5      &75.9       &78.5        &46.3   &59.1                      &58.4   &64.0   &70.7   \\
Intersection    &49.7      &72.9       &77.1        &42.9   &59.1                      &52.0   &59.3   &69.7   \\
M3S             &61.5      &75.6       &77.8        &56.1   &62.1                      &59.2   &64.4   &70.6   \\
GCN             &42.6      &67.8       &74.9        &33.4   &46.5                      &46.4   &49.7   &56.3   \\
Linear Snowball &69.5      &79.4       &80.4        &56.8   &65.4                      &64.1   &69.5   &72.9   \\
Snowball        &67.2      &78.5       &80.0        &56.4   &65.0                      &62.9   &68.3   &73.3   \\
Truncated Krylov&73.0      &{80.3}     &81.5        &59.6   &66.0                      &69.1   &{71.8} &{76.1} \\
VCHN            &{73.9}  &\textbf{81.3} &{82.0}     &{64.3} &{67.9}                    &{69.2} &{71.8} &75.2   \\
\midrule
\textbf{SACN} &\textbf{74.5}  &{78.3}   &\textbf{83.1}  &\textbf{66.9} &\textbf{68.4}    &\textbf{71.4} &\textbf{73.4} &\textbf{76.6}  \\
\bottomrule
\end{tabular*}
\end{table*}
\begin{table*}[th]
\centering
  \caption{Comparing to Contrastive Models, Accuracy (\textbf{Bold} values indicate the best).}
  \label{Contrastive}
  \begin{tabular*}{\textwidth}{@{\extracolsep{\fill}\quad}l|ccc|cc|ccc}
  \toprule
  Dataset        &\multicolumn{3}{c|}{Cora}       &\multicolumn{2}{c|}{Citeseer}  &\multicolumn{3}{c}{PubMed} \\
  Label rate     &0.5\%    &1\%        &3\%        &0.5\%  &1\%   &0.03\%        &0.05\%   &0.1\% \\
  \midrule
  DGI            &67.5     &72.4       &78.9       &60.7   &66.9   &60.2         &68.4      &70.7   \\
  GMI            &67.1     &71.0       &78.8       &56.2   &63.5   &60.1         &62.4      &71.4   \\
  MVGRL          &61.6     &65.2       &79.0       &61.7   &66.6   &63.3         &69.4      &72.2   \\
  GRACE          &60.4     &70.2       &75.8       &55.4   &59.3   &64.4         &67.5      &72.3   \\
  CG$^3$         &69.3     &74.1       &79.9       &62.7 &\textbf{70.6} &68.3    &70.1      &73.2   \\
  VCHN           &{73.9} &\textbf{81.3} &{82.0}    &{64.3} &67.9   &{69.2}       &{71.8}    &{75.2}  \\
  \midrule
  \textbf{SACN}  &\textbf{74.5}  &{78.3} &\textbf{83.1} &\textbf{66.9} &{68.4} &\textbf{71.4} &\textbf{73.4} &\textbf{76.6}  \\\bottomrule
  \end{tabular*}
  \end{table*}
\subsection{Experimental Setting}
The proposed SACN method is constructed by two attentional layers and consists of three views. For the two strongly augmented views, we perform random feature masking.
For each view, we set the dropout rate, weight decay and the size of the second hidden layer as 0.6, $5\times10^{-4}$ and 16, separately. We train SACN end-to-end and the learning rate is set to 0.01.
Following \cite{luan-et-al:multiscale},  we split the data into one small sample subset for training, a validation sample subset with 500 samples and a rest one with 1000 samples. 
As shown in Algorithm~\ref{algorithm}, we pretrain SACN for 100 epochs with the loss Eq.~\eqref{loss_sup} on labeled nodes, then train SACN with the overall loss Eq.~\eqref{loss_overall} on the unlabeled nodes. 
With the converge process, we add more pseudolabels for training. 
Moreover, to speed up the convergence and improve the test accuracy, we use a renormalization filter \cite{li-et-al:graphfiltering} to preprocess node features in the first stage.   
Let $\widehat{A}$ represents the adjacency matrix been symmetrically normalized, formalize as $\widehat{A} = \widetilde{D}^{(-1/2)}\widetilde{A}\widetilde{D}^{(-1/2)}$, where $\widetilde{A} = A + I_{n}$, $\widetilde{D}$ is the degree matrix of $\widetilde{A}$ and $I_{n}$ denotes the identity matrix. 
Then, we normalize the features by $X \leftarrow \widehat{A}^cX$, where $c$ represents the adjustable filtering strength.
 Low label rates correspond to high filtering strengths.
 For different label rates, we set different filtering strengths $c$ varying from 15 to 3.  
 In order to reduce the accidental error, we conduct 10 runs using the optimal hyper parameters for each experiment, and use the average accuracy as the final result.
\begin{figure*}[th]
  \centering
  \includegraphics[width=1.0\textwidth]{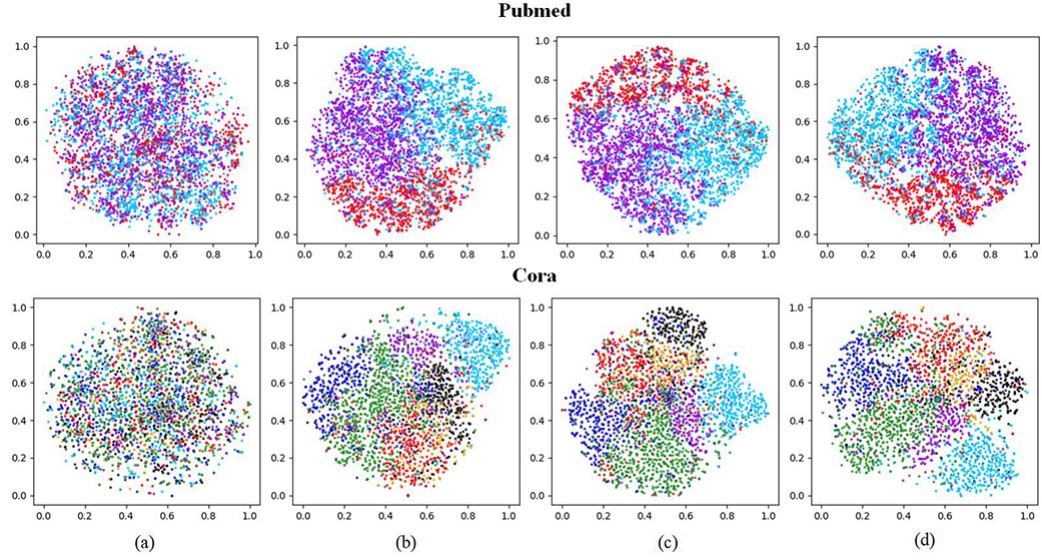}\\
\caption{$t$-SNE visualization of the learned representations from SACN algorithm on Coar, Citeseer and Pubmed separately during training, different categories of points are colored differently. In the above figures: (a) The input data X. (b) Feature after 100 epochs.  (c) Feature after 200 epochs.  (d) Feature after 400 epochs. With the increase of epochs, representations of the same class are close while those of different classes are far away gradually. }
\label{fig03}
\end{figure*}

\subsection{Comparison}
We compare the proposed SACN approach with several baseline methods, \eg,
Chebyshev networks~\cite{dbv:NeurIPS2016}, GCN for fingerprint (GCN-FP)~\cite{duvenaudt-et-al:fingerprints}, graph sample and aggregate (GraphSAGE)~\cite{hyl:inductive}, Graph Attention neTworks (GAT)~\cite{graph2018}, Graph Convolutional Networks (GCN)~\cite{kw:semiGCN}, LanczosNet (LNet)~\cite{liao:LanczosNet}, AdaLanczosNet (AdaLNet)~\cite{liao:LanczosNet}, linear snowball GCN, snowball GCN and truncated block Krylov network in~\cite{luan-et-al:multiscale}, and the heterogeneous model VCHN~\cite{liao-et-al:View-Consis} based on GCN and GAT.
We also compare it with methods that do not use validation, including label propagation using Partial Absorption Random Walk (PARW)  (LP)~\cite{wu-et-al:PartialAbsorbRW}, Co-training~\cite{lhw:AAAI2018}, Self-training~\cite{lhw:AAAI2018}, Union~\cite{lhw:AAAI2018}, Intersection~\cite{lhw:AAAI2018}, Multi-stage self-supervised training and M3S\cite{slz:multi2019}.

We compare SACN to the following state-of-the-art methods: LP~\cite{wu-et-al:PartialAbsorbRW} introduces the newly proposed PARW into label propagation. Chebshev~\cite{dbv:NeurIPS2016} uses $K$-th order Chebyshev filter to perform convolutions. GCN~\cite{kw:semiGCN} Computes localized first-order approximation of spectral graph convolution. Co-training~\cite{lhw:AAAI2018} employs a random walk model first to explore graph structure then train GCN. Self-training~\cite{lhw:AAAI2018} repeats to train GCN and select the most confident labels until meeting the conditions. Union and Intersection~\cite{lhw:AAAI2018} add the union or intersection of confident predictions from co-training and self-training. LNet~\cite{liao:LanczosNet} approximates the Laplacian matrix of graph using Lanczos algorithm with fixed structure. AdaLNet~\cite{liao:LanczosNet} adapts to different graph kernel function and node embeddings using backpropagation. M3S ~\cite{slz:multi2019} trains with multi stage and adds the most confident nodes at each stage, the latter uses Deep-cluster to choose nodes additionally. Snowball~\cite{luan-et-al:multiscale} expands the range of neighbors used by each node with the increase of the network layers. Linear Snowball~\cite{luan-et-al:multiscale} adopts snowball architecture but uses linear function as activation. Truncated Krylov~\cite{luan-et-al:multiscale} merges GCN into the block Krylov framework by truncation. VCHN~\cite{liao-et-al:View-Consis} uses heterogeneous networks and view-consistency loss to optimize.

Tables~\ref{with} and \ref{without} show the test accuracy of SACN and the above mentioned methods in different datasets and different label rates. 
Concretely, for Cora and Citeseer, when the label rates are 0.5\% and 1\% there are only two and three labeled nodes per class. For Pubmed, 0.03\% indicates it merely have two labeled nodes for each class. The results from other baseline methods are directly excerpted from the original papers. SACN outperforms the baselines in most cases especially when the label rates are very low. 
Differing from the traditional single network methods, SACN carries out different forms of information sharing and complementary across views. We think this is the main reason that our approach outperforms the state-of-the-art Truncated Krylov network on Citeseer without setting the validation set. 

Table~\ref{Contrastive} shows the results comparing to contrastive methods: DGI~\cite{DGI}, GMI~\cite{peng-et-al:MutualInfoMax}, MVGRL~\cite{hk:contramulti}, CG$^3$~\cite{wan-et-al:Contrastive2021}, GRACE~\cite{zhu-et-al:deep2020} and also VCHN.

We compare SACN to the following contrastive methods:  DGI~\cite{DGI} introduces mutual information maximization into graph structure. GMI~\cite{peng-et-al:MutualInfoMax} improves DGI to maximize mutual information from the coarse-grained graph level to the fine-grained node level. MVGRL~\cite{hk:contramulti} augments the graph structure to achieve infomax between augmented views. CG$^3$~\cite{wan-et-al:Contrastive2021} combines the contrastive loss of local features with the generative loss of global features from GCN. GRACE~\cite{zhu-et-al:deep2020} mainly focuses on the comparison of local node representations.

Besides, we utilize the $t$-SNE algorithm ~\cite{lg:tSNE} to visualize the embeddings learned on Cora during training. As shown in Fig.~\ref{fig03}, it is 
straightforward to check that representations of nodes from the same class become closer, while the boundaries between different classes become clearer with the training process.

\begin{table}[H]
\centering
\caption{Comparison of different pseudolabel learning methods.}
\label{C}
\begin{tabular*}{0.46\textwidth}{@{\extracolsep{\fill}\quad}l|ccc}
\toprule
Method                 &Cora        &Citeseer       &PubMed \\
\midrule
Self-training          &53.7        &43.3           &51.9    \\
Using auxiliary task   &58.5        &46.3           &58.4    \\
VCHN                   &74.9        &65.6           &71.8    \\
\midrule
SACN               &\textbf{76.2}  &\textbf{67.0}   &\textbf{72.2}  \\
\bottomrule
\end{tabular*}
\end{table}
In Table~\ref{C}, the first three methods correspond to the three existing pseudolabel generation methods mentioned in Fig.~\ref{fig01}. SACN surpasses these three methods and shows competitiveness for the node classification task.
\subsection {Computational Complexity and Parameter Counts}
The following table~\ref{counts} compares the network parameters of SACN with previous contrastive methods. It is evident from the table that the counts of model parameters in SACN is reduced by at least one-third compared to other contrastive models. This reflects the competitiveness of SACN in terms of occupying less memory, making it more friendly when processing large-scale graphs and improving efficiency. The integration of structural information also reduces the computational complexity to some extent. We know that the complexity of matrix multiplication is $O(n^3)$, but in SACN, the integration of neighborhood will reduce the real computational complexity to be lower than $O(n^3)$ for the sparse structure of graph that most nodes connect with only few other nodes. 
\begin{table}[H]
\centering
\caption{Comparison of Parameter Counts on different pseudolabel learning methods.}
\label{counts}
\begin{tabular*}{0.26\textwidth}{@{\extracolsep{\fill}
\quad}l|r}
\toprule
Method                 &Parameter Counts         \\
\midrule
DGI~\cite{DGI}          &99,364       \\
GMI~\cite{peng-et-al:MutualInfoMax}   &1,992,709        \\
MVGRL~\cite{hk:contramulti}   &1,730,563        \\
GRACE~\cite{zhu-et-al:deep2020}   &1,102,592        \\
VCHN~\cite{liao-et-al:View-Consis}   &92,517        \\
\midrule
SACN               &\textbf{69,230} \\
\bottomrule
\end{tabular*}
\end{table}

\subsection{Ablation Study}
We conduct ablation experiments to verify the usefulness of each component in SACN. The results of the ablation experiment are shown in Table~\ref{A}. By removing $\ell_{\rm sup}$, $\ell_{\rm  sacn}$, and $\ell_{\rm  w2s}$ separately, under the same parameter settings, we observe a decline in the classification performance of model. This finding clearly demonstrates the significant improvements brought by SACN through its structure-aware consensus loss and weak-to-strong loss, further proving the reliability of the pseudo-labels utilized in these strategies.
\begin{table}[H]
\centering
\caption{Ablation study using different loss terms.}
\label{A}
\begin{tabular*}{0.46\textwidth}{@{\extracolsep{\fill}
\quad}ccc|ccc}
\toprule
$\ell_{\rm sup}$ &$\ell_{\rm  sacn}$  &$\ell_{\rm  w2s}$  &Cora   &Citeseer  &PubMed \\
\midrule
$\surd$     &          &$\surd$        &74.2   &65.8      &66.4    \\
$\surd$     &$\surd$     &             &70.6   &55.3      &64.9     \\
$\surd$     &$\surd$     &$\surd$  &\textbf{76.2} &\textbf{67.0}  &\textbf{72.2} \\
\bottomrule
\end{tabular*}
\end{table}

By combining view consistency and pseudolabel strategy, we make full use of unlabeled nodes and integrate useful information of different views. These experiments are conducted on three datasets with two labeled nodes.
\section{Conclusion}\label{Conclusions}
Existing methods for semi-supervised node classification primarily rely on labeled samples. However, labeled samples are typically limited in most datasets, and the datasets themselves suffer from class imbalance. Therefore, performing semi-supervised node classification with a very limited number of labeled nodes is a very challenging task.

To address these limitations, we propose Structure-Aware Consensus Network. We draw on the consensus learning between strongly augmented views, meanwhile integrate weak-to-strong pseudo supervision. 
For present semi-supervised methods designed to handle the limited labels, either pseudolabels are underutilized as supervision or depend on the complex design of heterogeneous networks. 
Different from them, the proposed SACN achieves consensus on nodes from different augmented views while maintaining a single network. This strong-to-strong consensus learning innovatively incorporates the neighborhood information unique to graph structure. Considering the common insufficiency of labels, we introduce pseudo supervision from weakly augmented view to strong ones. 
What makes SACN being more distinguishable is that, the pseudolabels are selected on each class, namely the class-aware strategy, it is proposed to solve the mentioned class imbalance. 
In general, the improvement of SACN is attributed to three novelties. One is the creatively combination of weak-to-strong supervision and strong-to-strong consensus learning within a single net work. This creativity not only handles the realistic of the limited labeled data but also integrate unique structural correlation to make improvements. 
The second one is the new structure-aware consensus objective distinguishing itself from other CCA-like methods due to its computational simplicity. 
Lastly, the challenge of class imbalance is also addressed by the implementation of the class-aware pseudolabel selection. An connotative insight for above is, the latter two make SACN more applicable for very large graphs, for the generality of class imbalance among large ones and the low requirements for computation of Eq.~\eqref{inv_loss}. 

The experiments conduct on three widely-used graph datasets demonstrate the effectiveness of SACN even when label rates are very low. Specifically, in the node classification task, SACN consistently achieves state-of-the-art performance, irrespective of whether the validation set is utilized.

\bibliography{ijcai24}
\bibliographystyle{IEEEtran}

\end{document}